\pdfoutput=1

\documentclass[11pt]{article}

\usepackage[]{acl}

\usepackage{times}
\usepackage{latexsym}
\usepackage{graphicx}
\usepackage{hyperref}       
\usepackage{tabularx}

\usepackage[T1]{fontenc}

\usepackage[utf8]{inputenc}

\usepackage{microtype}

\usepackage{inconsolata}
\usepackage{subcaption}
\usepackage{tablefootnote}
\usepackage{soul}
\usepackage{booktabs}
\usepackage{tikz}
\usepackage[tikz]{bclogo}
\usepackage{pgfplotstable}
\usepackage{mdframed}
\usepackage{graphicx}
\usepackage{amsmath}
\usepackage{algorithm}
\usepackage{algpseudocode}
\usepackage{wrapfig}
\usepackage[fixed]{fontawesome5}
\usepackage{multirow}
\usepackage{makecell}
\usepackage[noabbrev,capitalize]{cleveref}

\usepackage{colortbl}
\newcommand{\headercolor}{\rowcolor{gray!15}}
\newcommand{\ours}{{\sc{AdaSwitch}}}
\newcommand{\methodname}{{\sc{AdaSwitch}}}
\newcommand{\oursbf}{{\sc{\textbf{AdaSwitch}}}}

\title{{\sc{\textbf{AdaSwitch}}}: Adaptive Switching between Small and Large Agents for Effective Cloud-Local Collaborative Learning}

\author{
Hao Sun\textsuperscript{1,2}, 
Jiayi Wu\textsuperscript{3}, 
Hengyi Cai\textsuperscript{4}, 
Xiaochi Wei\textsuperscript{5} \\
\textbf{Yue Feng\textsuperscript{7}, 
Bo Wang\textsuperscript{6}, 
Shuaiqiang Wang\textsuperscript{5}, 
Yan Zhang\textsuperscript{1,2}, 
Dawei Yin\textsuperscript{5}}
\\
\textsuperscript{1}State Key Laboratory of General Artificial Intelligence, Peking University, Beijing, China\\
\textsuperscript{2}School of Intelligence Science and Technology, Peking University\\
\textsuperscript{3}East China Normal University,
\textsuperscript{4}Chinese Academy of Sciences\\
\textsuperscript{5}Baidu Inc, 
\textsuperscript{6}Beijing Institute of Technology,
\textsuperscript{7}University of Birmingham\\
\tt{sunhao@stu.pku.edu.cn}\\
}

\begin{document}
\maketitle
\begin{abstract}
Recent advancements in large language models (LLMs) have been remarkable.
Users face a choice between using cloud-based LLMs for generation quality and deploying local-based LLMs for lower computational cost.
The former option is typically costly and inefficient, while the latter usually fails to deliver satisfactory performance for reasoning steps requiring deliberate thought processes.
In this work, we propose a novel LLM utilization paradigm that facilitates the collaborative operation of large cloud-based LLMs and smaller local-deployed LLMs.
Our framework comprises two primary modules: the local agent instantiated with a relatively smaller LLM, handling less complex reasoning steps, and the cloud agent equipped with a larger LLM, managing more intricate reasoning steps.
This collaborative processing is enabled through an adaptive mechanism where the local agent introspectively identifies errors and proactively seeks assistance from the cloud agent, thereby effectively integrating the strengths of both locally-deployed and cloud-based LLMs, resulting in significant enhancements in task completion performance and efficiency.
We evaluate \methodname{} across 7 benchmarks, ranging from mathematical reasoning and complex question answering, using various types of LLMs to instantiate the local and cloud agents.
The empirical results show that \methodname{} effectively improves the performance of the local agent, and sometimes achieves competitive results compared to the cloud agent while utilizing much less computational overhead.
\end{abstract}

\section{Introduction}

Recently, the advent of large language models (LLMs) has garnered substantial attention from the public, industry, and academia, attributed to their advanced language comprehension and generation capabilities.
These LLMs, such as OpenAI's GPT-4~\citep{achiam2023gpt} and Google's PaLM~\citep{anil2023palm}, are characterized by their massive scale, both in terms of the colossal number of parameters and the substantial volume of data utilized during their training process.
Due to their large number of parameters, LLMs are typically deployed on cloud servers. 
However, the reliance on cloud computing makes LLM utilization considerably costly and inefficient, owing to the substantial bandwidth consumption, considerable strain on the network architecture and the need for extensive computational resources.

\begin{figure}[t]
  \centering
  \includegraphics[width=1.0\columnwidth]{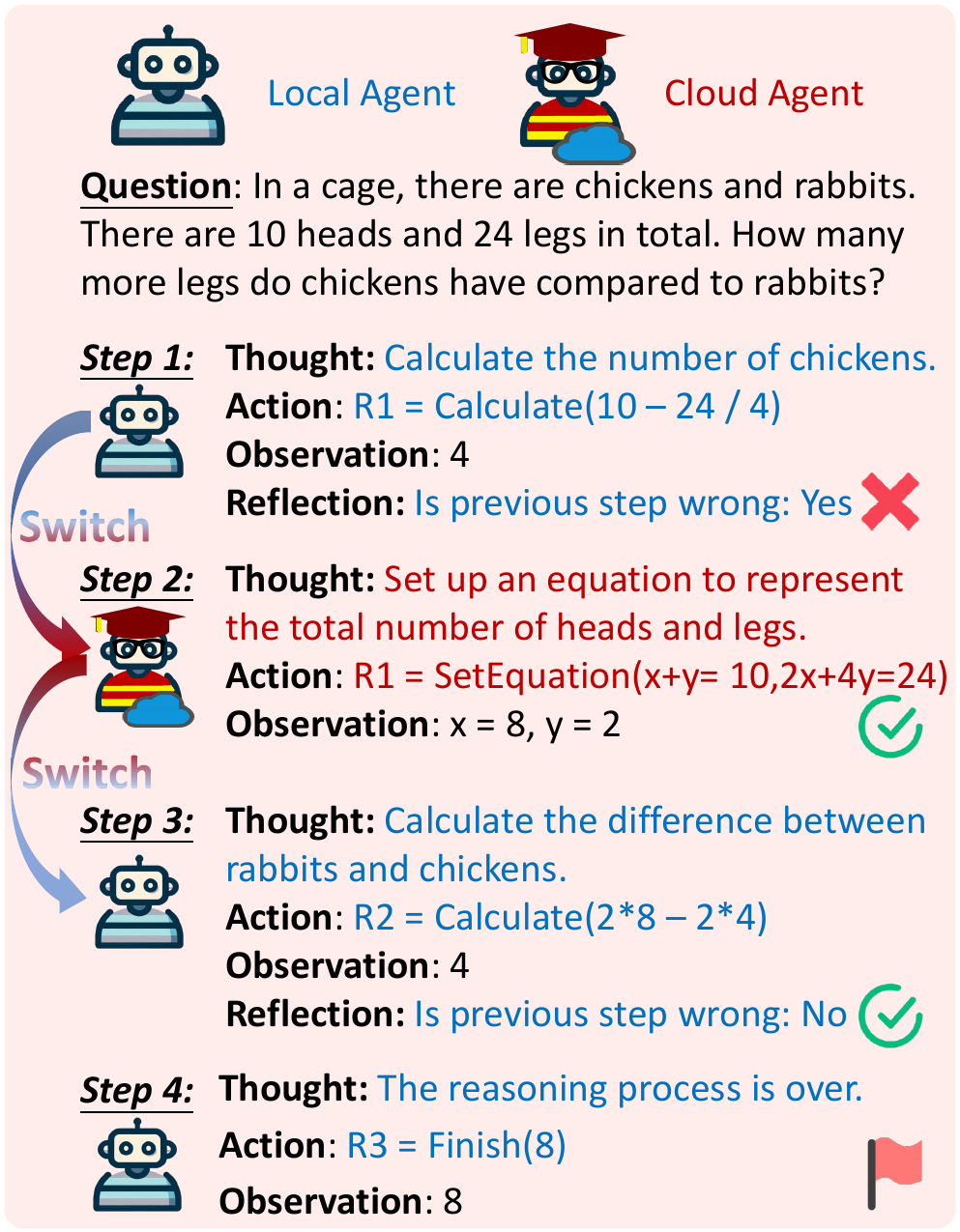}
      \caption{A brief illustration of \methodname{} framework, in which local agent and cloud agent alternate to collaboratively fulfill the given question.}
  \label{fig:introduction}
\end{figure}

A promising solution is to reduce the high computational demand of LLMs through techniques such as knowledge distillation~\citep{liang2020mixkd,gu2023minillm} or model quantization~\citep{frantar2022gptq,lin2023awq,xiao2023smoothquant}, and to deploy LLMs directly on local devices.
Though effective, small-sized LLMs are prone to severe performance degradation when confronted with complex and demanding situations, and usually fail to deliver satisfactory performance for reasoning steps requiring deliberate thought processes.

To harness the strong capabilities of large-sized LLMs with the convenience of small-sized LLMs, we propose \methodname{}, a novel framework enabling these two types of LLMs to collaboratively solve complex open-world tasks.
This framework is inspired by human behavior in similar scenarios: when faced with complex tasks, people often seek assistance from more knowledgeable individuals for challenging components and learn from their guidance to complete the tasks. This ability to proactively seek assistance and apply acquired knowledge is a critical aspect of human intelligence.
Similarly, \methodname{} comprises two primary modules: the local agent and the cloud agent. The local agent, instantiated with a relatively smaller LLM, is capable of handling less complex reasoning steps. 
In contrast, the cloud agent, responsible for more deliberate reasoning, utilizes larger LLMs, such as \texttt{Llama-33B}.
Our proposed approach is designed to enable both efficient inference with local smaller LLMs and resource-intensive cloud LLM executions for task steps requiring higher cognitive capabilities.
It effectively integrates the strengths of both locally-deployed and cloud-based LLMs, resulting in significant enhancements in task completion performance and efficiency.

As depicted in Figure~\ref{fig:introduction}, given the question ``how many more legs do chickens have compared to rabbits?'', \methodname{} interleaves its generation by first composing the sub-solution ``calculate the number of chickens'', then reflecting on the prior failing step, and offloading this challenging step to the cloud agent.
The local agent thus is able to form an improved action to finally accomplish the task.
Using the cloud agent as an assistant allows the local agent to make effective use of a larger knowledge base and focus its efforts on learning the task steps appropriate for the model's current competence.

The main idea behind \methodname{} is to allow the local agent to adaptively activate the cloud agent when it introspectively judges the current step as incorrect. 
To this end, we enhance the local agent's self-checking capabilities by meticulously collecting inaccurate reasoning paths to construct the mistake-checking dataset.
Specifically, we ask the local agent to undertake an exam during which the cloud agent dynamically corrects the local agent's mistakes, incentivizing the local agent to learn from mistakes, determine when to ask for assistance, and how to utilize feedback to correct the mistake.
Finally, the resultant local agent can introspectively judge the running steps, proactively seek assistance, and apply acquired feedback to improve its subsequent actions.

We conduct experiments on mathematical reasoning and complex reasoning benchmarks.
\methodname{} consistently improves the performance across various LLMs and tasks.
For instance, the performance of the local model instantiated with the \texttt{DeepSeek-Coder-1.3B} can be improved from 29.3\% to 53.9\%, requiring 3x fewer computational costs for LLM inference than competitor system while achieving similar results.
Notably, our proposed framework even enables \texttt{StarCoder2-3B} to achieve comparable performance against \texttt{Llama-33B}, with 5x less computational overhead for LLM inference.
The effectiveness of the proposed method is also verified by ablation experiments and analytical experiments.
\section{Methodology}
\begin{figure*}[tb]
\includegraphics[width=\textwidth]{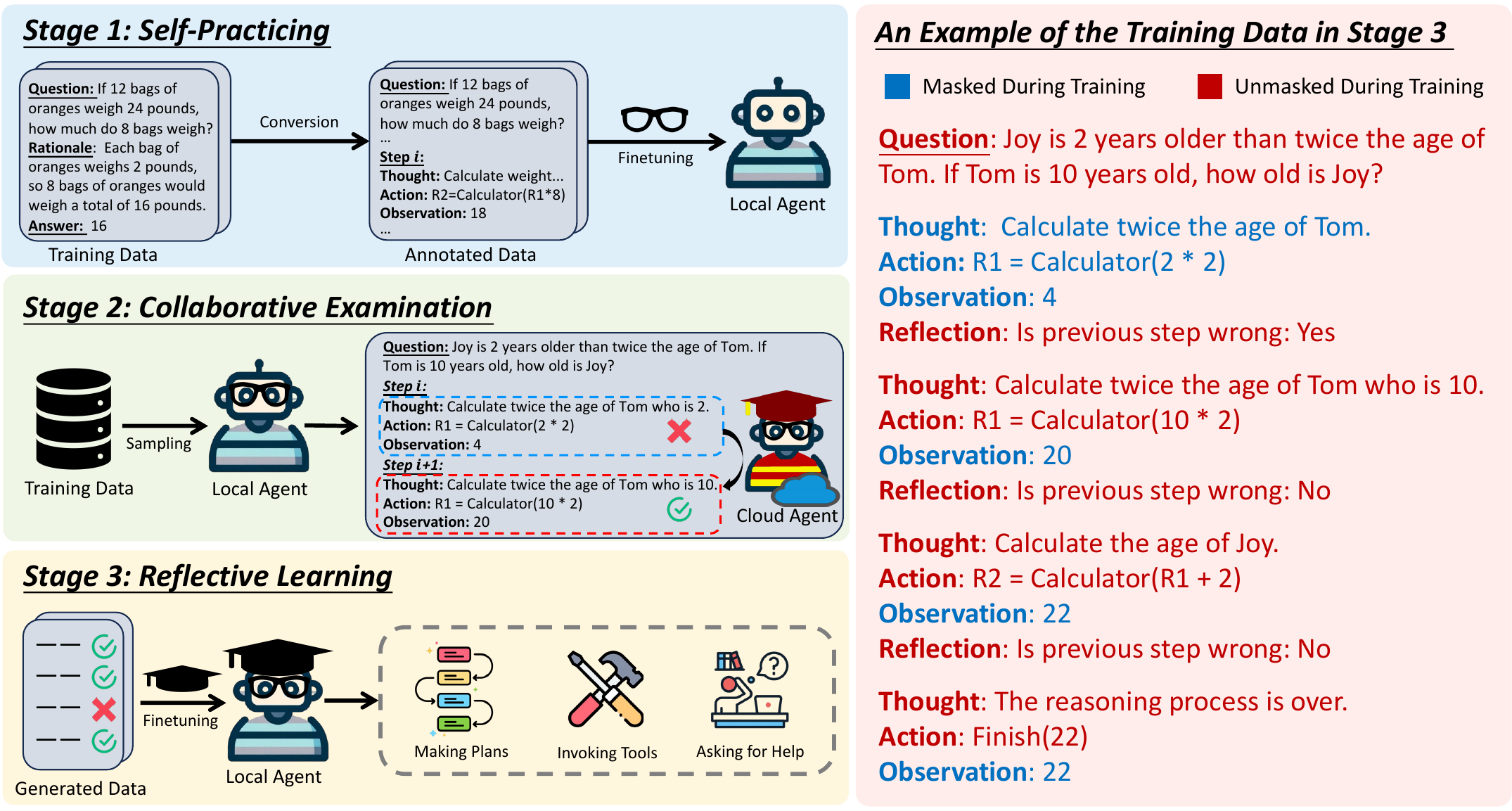}
\centering
\caption{The illustration of \ours{}. 1) Self-Practicing: The local agent practices on the training dataset to build the basic reasoning ability. 2) Collaborative Examination: The local agent undertakes an exam to expose its weakness, during which the cloud agent will be utilized to correct the mistakes. 3) Reflective Learning: The local agent is trained on the mistake-correction trajectories generated in the second stage.}
\label{fig:model}
\end{figure*}
\subsection{Preliminary}
In the agent framework, the agents usually follow the interaction paradigm, where the agent predicts a thought and an action, and the environment gives feedback.
Specifically, the backbone of the agent is an LLM denoted as $\mathcal{M}$. 
In the $t$-th step, the LLM $\mathcal{M}$ generates a thought $s_t$ and an action $a_t$ based on the instruction and the current state of the system:
\begin{align}
    s_t, a_t &= \mathcal{M}(\tau_{t-1}), \\
    o_t &= \text{Execution}(a_t)
\end{align}
where $\tau_{t-1}=\{s_1,a_1,o_1,...,s_{t-1},a_{t-1},$ $o_{t-1}\}$ denotes the previous interaction trajectory. Here, $o_{t}$ denotes the observation returned by tools when the action $a_t$ is executed.
The tool list used in this paper is shown in \cref{table:appendix-tools}.

\subsection{\oursbf{} Framework}
We propose a multi-stage learning paradigm that enables the local agent to introspectively judge the running steps, proactively seek assistance, and apply acquired feedback to improve its subsequent actions.
Specifically, as shown in Figure~\ref{fig:model}, the learning of our framework can be divided into three stages:
(1) firstly, the local agent is trained on the training set to build a basic reasoning ability;
(2) then, the local agent undertakes an exam to expose its weakness in accomplishing the challenging steps, during which the cloud agents are utilized to correct the local agents' mistakes;
(3) finally, the local agents are trained on the mistake-checking and mistake-correction trajectories generated in the second stage.

\paragraph{Self-Practicing}
Given a training dataset $D_{train} = \{ \langle x_i,r_i, y_i \rangle\}_{i=1}^{k}$, where $x_i$ denotes the question, $r_i$ represents the ground-truth intermediate reasoning steps and $y_i$ is the answer, we follow \citet{yin2023lumos} and employ LLM to transform the reasoning steps $r$ into interaction trajectories.
Specifically, we provide the LLM with the question, ground-truth intermediate reasoning steps, and defined action space, then the LLM is able to generate high-level thoughts and corresponding actions accordingly.
Subsequently, we can obtain the interaction trajectories in the format of $\tau=\{s_1,a_1,o_1,...,s_n,a_n,o_n\}$.
To enable the local agent to focus on the reasoning part, we feed the entire interaction trajectories to the LLMs while merely calculating the decoding loss on the tokens of the subgoals and action by applying binary masking on observation tokens.

\paragraph{Collaborative Examination}
After training on the annotated dataset, the local agent can solve the question interactively.
To detect the weakness of the local agent, we ask the local agent to undertake an exam on the training set $D_{train}$.
For each question, we collect up to four interaction trajectories through decoding with the Top-K sampling strategy.
During the interaction, we follow  \citet{li2022making} and adopt a rule-based method to dynamically examine whether each reasoning step is wrong.
For mathematical reasoning tasks, we gather intermediate results from the correct trajectories and verify whether each step’s execution result matches any of the intermediate results in the correct trajectories. If there is a match, the step is deemed correct; otherwise, it is considered incorrect.
In the case of textual reasoning tasks, we can similarly verify the correctness of each step by utilizing \texttt{roberta-large-mnli} \cite{liu2019roberta} to check whether the thought and action of each step are semantically equivalent to any of the reasoning steps in the correct trajectories.


When a mistake is detected, the cloud agent will be activated to correct the mistake by erasing the wrong step in the prompt and regenerating the step.
Then the local agent will continue based on the cloud agent's corrected step.
This process will continue until the local agent finally reaches the answer.
If the answer is true, we incorporate the interaction trajectory into the training set.

\paragraph{Reflective Learning}
After obtaining the mistake-checking and mistake-correction data, we ask the local agent to train on the revised trajectories.
It is worth mentioning that the revised trajectories contain wrong steps that have wrong thoughts and actions.
Therefore, we mask these parts during loss calculations so that the local agent will not be confused by these steps.

\paragraph{Collaborative Inference}
After training on the mistake-checking and mistake-correction dataset, the local agent can detect and correct errors.
We introduce two modes for task inference.
The first mode is {\sc{self-reflection}}, where the local agent will rely on itself to correct the mistake when it finds an error.
However, due to the limited ability of the local agent, it might not be able to correct the mistake accurately.
Therefore, we further introduce the second mode where the cloud agent will be activated to correct the mistake, which we call \ours{}.
Specifically, the local agent predicts a probability that the previous step is wrong, and the cloud agent will be activated when the probability is larger than an activation threshold $p$.
\section{Experiment}
\begin{table*}[tbp]
\centering
\resizebox{1.0\textwidth}{!}{
\begin{tabular}{l|c|cccccccc}
\toprule
\multicolumn{1}{c|}{\multirow{3}[4]{*}{\textbf{Method}}} & \multirow{3}[4]{*}{\textbf{\# Para}} & \multicolumn{5}{c}{\textbf{Mathematical Reasoning}} & \multicolumn{3}{c}{\textbf{Complex QA Reasoning}} \\
\cmidrule(lr){3-7} \cmidrule(lr){8-10}
&       & \textbf{GSM8K} & \textbf{G\_Hard} & \textbf{SVAMP} & \textbf{ASDIV} &  \textbf{MultiArith} & \textbf{MuSiQue} & \textbf{HotpotQA} \\
\midrule
\headercolor
\multicolumn{9}{c}{\textbf{Using 1.3B Local Agent}} \\
\textbf{Local Agent} & 1.3B    & 29.30  & 25.20  & 26.60  & 43.90  & 77.22 & 29.80 & 25.80 \\
\textbf{+\methodname{}} & 1.3B    & 53.90 (+24.6)  & 47.10 (+21.9)  & 46.90 (+20.3)  & 61.90 (+18.0)  & 94.44 (+17.2)  & 36.80 (+7.0) & 32.50 (+6.7) \\
\midrule
\headercolor
\multicolumn{9}{c}{\textbf{Using 3B Local Agent}} \\
\textbf{Local Agent} & 3B      & 48.80  & 40.10  & 37.80  & 52.50  & 87.22  & 31.50 & 29.50\\
\textbf{+\methodname{}} & 3B      & 60.60 (+11.8)  & 50.60 (+10.5)  & 52.60 (+14.8)  & 66.20 (+13.7)  & 96.11 (+8.9)  & 37.80 (+6.3) & 31.80 (+2.3)\\
\midrule
\textbf{Cloud Agent} & 33B   & 63.20  & 55.00  & 52.10  & 63.80  & 98.89 &  41.80 &  35.50 \\
\bottomrule
\end{tabular}
}
\caption{Our main experimental results (\%) on five mathematical reasoning tasks and two complex question-answering tasks. Local agent refers to the agent after the self-practicing while \ours{} refers to undergoing the full learning paradigm and then collaborating with the cloud agent during inference.}
\label{tab:main}%
\end{table*}%

\subsection{Tasks \& Datasets}

\paragraph{Mathematical Task}
We adopt five math word problem datasets to evaluate the mathematical reasoning ability.
GSM8K is a primary school-level mathematical dataset~\citep{cobbe2021training}.
G\_Hard is a harder version of GSM8K~\citep{gao2022pal}.
MultiArith is a multi-step arithmetic reasoning dataset ~\citep{roy2016solving}.
SVAMP is created by applying chosen variations over examples sampled from existing datasets~\citep{patel2021nlp}.
ASDIV is a math word problem dataset that contains examples with diverse language patterns and problem types~\citep{miao2021diverse}.

\paragraph{Complex QA Task}
We use two open-domain question-answering datasets to evaluate the complex reasoning ability.
HotpotQA dataset~\citep{yang2018hotpotqa} is a multi-hop question-answering dataset.
MuSiQue dataset~\citep{trivedi2022musique} is a multi-hop reasoning dataset.

\subsection{Experimental Setup}
For mathematical tasks, we use GSM8K as the training dataset. For the complex QA task, we use MuSiQue as the training dataset.
The detailed statistics of these datasets are shown in Table~\ref{tab:datasets}.
Moreover, the cloud agent undergoes only the self-practicing stage to build basic reasoning ability before deployment, while the local agent undergoes the full learning stages.
\subsection{Models}
The candidate local agents include
\texttt{DeepSeek-Coder-1.3B}\footnote{\tiny \url{https://huggingface.co/deepseek-ai/deepseek-coder-1.3b-instruct}}
and \texttt{StarCoder2-3B}\footnote{\tiny \url{https://huggingface.co/bigcode/starcoder2-3b}}.
And the candidate cloud agents include
\texttt{CodeLlama-13B} \footnote{\tiny \url{https://huggingface.co/codellama/CodeLlama-13b-hf}},
\texttt{Llama-33B} \footnote{\tiny \url{https://huggingface.co/huggyllama/llama-30b}},
\texttt{Qwen1.5-32B} \footnote{\tiny \url{https://huggingface.co/Qwen/Qwen1.5-32B-Chat}},
and \texttt{Llama-2-70B} \footnote{\tiny \url{https://huggingface.co/meta-llama/Llama-2-70b-hf}}.

\subsection{Main Results}
In this section, we conduct experiments on seven challenging reasoning tasks utilizing the 1.3B and 3B local agents and the 33B cloud agent. The result is shown in \cref{tab:main}.
Based on the result, several observations can be made:

First, \textbf{our method, {\sc{\textbf{AdaSwitch}}}, greatly improves the performance of the local agent}, achieving up to 86.9\% relative improvement using the 1.3B model as the local agent and up to 39.1\% relative improvement using the 3B model as the local agent. This is primarily because, after collaborative learning, \ours{} enables the local agent to seek help from the cloud agent when it detects potential mistakes. By doing so, the local agent can handle the easier steps independently while leveraging the cloud agent for more difficult steps, thereby enhancing its overall performance.

Second, \textbf{the improvement is more pronounced on difficult datasets}, as evidenced by an 86.9\% relative improvement on G\_Hard dataset and a 22.3\% relative improvement on MultiArith dataset. This is mainly because the local agent learns to request assistance when necessary, and when faced with more challenging datasets, the local agent will call for help more frequently, leading to significant performance gains. However, the cost associated with these difficult datasets will also increase. To balance between cost and effectiveness, we can adjust the predefined activation threshold $p$, which is discussed in \cref{tab:hyper}.

\subsection{Ablation Study}
\begin{figure}[tb]
\includegraphics[width=0.95\linewidth]{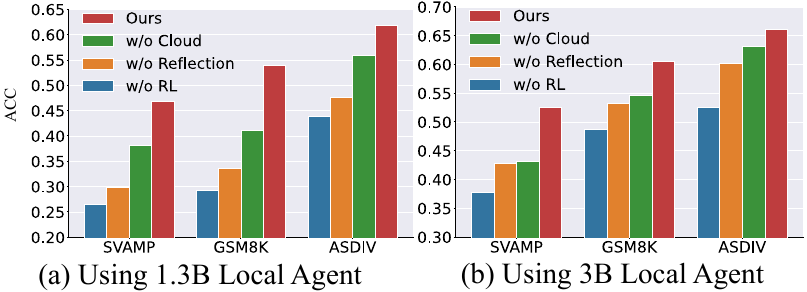}
\centering
\caption{We conduct an ablation study by removing the cloud agent, self-reflection, and reflective learning.}
\label{fig:ablation}
\end{figure}

In this section, we assess the performance of different inference modes using \texttt{DeepSeek-Coder-1.3B} and \texttt{StarCoder2-3B} as local agents, which is shown in \cref{fig:ablation}.
Specifically, w/o cloud agent refers to utilizing the local agent to self-correct the mistakes without asking for help from the cloud agent.
w/o reflection refers to continuing the inference processes without identifying or correcting errors in previous steps.
w/o RL refers to removing the reflective learning.

As we can see, after removing the cloud agent, the performance degrades dramatically, which is reasonable because the local agent cannot solve the difficult steps due to its limited reasoning ability.
However, by conducting self-reflection, the performance of w/o cloud is still higher than the performance of w/o reflection, which demonstrates that the local agent can correctly detect and correct mistakes on its own.
Moreover, the performance of w/o reflection is higher than w/o RL, demonstrating that after reflective learning, the local agent's reasoning ability gets improved.

\subsection{Hyper-parameter Analysis}
\begin{table}[tbp]
\centering
\resizebox{1.0\linewidth}{!}{
\begin{tabular}{r|cccccc}
\toprule
 & \multicolumn{2}{c}{\textbf{GSM8K}} & \multicolumn{2}{c}{\textbf{SVAMP}} & \multicolumn{2}{c}{\textbf{ASDIV}} \\
 \cmidrule(lr){2-3} \cmidrule(lr){4-5} \cmidrule(lr){6-7}
\textbf{$p$} & \textbf{Acc} & \textbf{Cost} &\textbf{Acc} & \textbf{Cost} &\textbf{Acc} & \textbf{Cost} \\
\midrule
\textbf{0.1}   & 57.60 & 121.80 & 62.40 & 75.53 & 69.30 & 49.39  \\
\textbf{0.3}   & 57.30 & 62.95  & 58.40 & 53.21 & 67.40 & 36.39   \\
\textbf{0.5}   &  57.40 & 77.61 & 55.40 & 46.48 & 66.60 & 31.59  \\
\textbf{0.7}   & 53.10 & 49.18 & 51.60 & 29.93 & 65.50 & 26.70  \\
\textbf{0.9}   &  48.50 & 37.90 & 45.20 & 21.07 & 63.40 & 23.49 \\
\bottomrule
\end{tabular}
}
\caption{Results (\%) of \ours{} using different activation threshold. As the threshold increases, \ours{} activates the cloud agent more frequently, leading to improved performance but at a higher inference cost. The unit of the cost is TFLOPs.}
\label{tab:hyper}%
\end{table}%

In our method, \ours{}, the decision for the local agent to seek assistance from the cloud agent is controlled by an activation threshold, denoted as $p$. To evaluate the impact of different threshold values, we conducted experiments using $p$ values of \{0.1, 0.3, 0.5, 0.7, 0.9\} with a 1.3B local agent and a 70B cloud agent on the subset of three mathematical reasoning datasets.

As illustrated in \cref{tab:hyper}, lowering the threshold $p$ results in the local agent requesting help more frequently, which generally leads to enhanced model performance. However, this increased reliance on the cloud agent also incurs higher computational costs. For instance, on the GSM8K dataset, the cost escalates from 37.90 TFLOPs to 121.80 TFLOPs as the threshold decreases from 0.9 to 0.1.
Moreover, we observe a saturation effect in performance improvement as the threshold value is further reduced. While a decrease in threshold from 0.9 to 0.7 results in a 9.4\% increase in accuracy, a further reduction from 0.3 to 0.1 yields only a 0.5\% improvement. This diminishing return is primarily because, after a certain point, the local agent’s performance closely approximates that of the cloud agent. Consequently, further reducing the threshold may result in the local agent seeking assistance on steps where the cloud agent is also prone to errors, thereby minimally impacting overall accuracy.

\subsection{Analysis}
\paragraph{Switching Analysis}
To demonstrate the effectiveness of various switching strategies, we compare it with the following variants:
Random Switch: The local agent switches to the cloud agent with a probability of $p$.
Sequential Switch: The local agent switches to the cloud agent every consecutive $k$ steps.
Confidence Switch: The local agent switches to the cloud agent when the probability of its generated tokens is lower than a predefined threshold $p$.
To ensure a fair comparison, we tune these corresponding hyperparameters to keep all methods at a similar cost level.

As shown in \cref{tab:switching}, all methods improve the performance of the local agent after collaborating with the cloud agent.
Among the variants, the Confidence Switch improves the performance higher than others.
This is mainly because the token probability distribution can reflect the local agent's capability in solving the question to some extent, making the cost quota be distributed properly.
However, the overconfidence phenomenon \cite{yang2024can, groot2024overconfidence, xiong2023can} makes the probability distribution an unstable metric, leading to inferior performance compared with our method.

\paragraph{Capability of Self-checking}
During the collaborative examination, we assess the correctness of each step using predefined rules. We first analyze the accuracy of the labeling process and then analyze if the local agent can correctly predict label accuracy.
We randomly selected 100 questions and manually checked the rule-based labels. Each step’s ground truth label is positive if correct and negative otherwise. The True Positive Rate (TPR) is the proportion of correct steps identified correctly, and the True Negative Rate (TNR) is the proportion of incorrect steps identified correctly. Our analysis showed that the rule-based method achieved a TPR of 92\% and a TNR of 61\%. The high TPR indicates the rule labeling process effectively identifies correct steps, while the low TNR suggests some steps are wrong though the result of the step has appeared in the correct trajectories. Such corner cases need human intervention to further verify the correctness of each step.

To verify if the local agent can learn to predict the correctness of each step, we randomly selected 100 inference trajectories of \ours{} and Confidence Switch. Our analysis showed that \ours{} achieved a TPR of 82\% and a TNR of 52\%, while Confidence Switch achieved a TPR of 73\% and a TNR of 27\%. These results confirm the effectiveness of our method. Notably, \ours{}’s TPR and TNR closely match the label data, showing it effectively utilizes training data to improve its ability to detect mistakes. Higher performance could be achieved with better training data, such as using LLMs as data labelers, which is suggested for future work.

\begin{table}[tbp]
\centering
\resizebox{1.0\linewidth}{!}{
\begin{tabular}{r|cccc}
\toprule
\textbf{Method} & \textbf{GSM8K} & \textbf{G\_Hard} & \textbf{SVAMP} &  \textbf{MultiArith} 	 \\
\midrule

\textbf{13B}   & 61.70 & 57.10 & 49.80  & 98.89 \\
\textbf{32B}   & 70.40 & 59.80 & 67.90  & 100.00 \\
\textbf{70B}   & 74.90 & 64.00 & 74.70 & 97.78 \\
\midrule

\headercolor
\multicolumn{5}{c}{\textbf{Using 1.3B Local Agent}} \\
\textbf{Local}  & 29.30 & 25.20 & 26.60 & 77.22 \\
\textbf{ + 13B}  & 54.24 & 48.89 & 44.58 & 97.51 \\
\textbf{ + 32B}  &  58.80 & 51.60 & 52.67 & 96.65 \\
\textbf{ + 70B}  &  58.96 & 52.51 & 55.81 & 97.92 \\
\midrule
\headercolor
\multicolumn{5}{c}{\textbf{Using 3B Local Agent}} \\
\textbf{Local}  &  48.80 & 40.10 & 37.80 & 87.22 \\
\textbf{ + 13B}  &  61.23 & 49.95 & 50.30 & 97.56 \\
\textbf{ + 32B}  &   64.49 & 52.40 & 59.14 & 98.67 \\
\textbf{ + 70B}  &  65.43 & 53.66 & 61.36 & 98.28 \\
\bottomrule
\end{tabular}
}
\caption{Results (\%)  using different cloud agent models, where +13B refers to collaborating the local agent with the 13B cloud agent.}
\label{tab:general}%
\end{table}%
\begin{table*}[tbp]
\centering
\resizebox{1.0\textwidth}{!}{
\begin{tabular}{l|c|cccccccc}
\toprule
\multicolumn{1}{c|}{\multirow{3}[4]{*}{\textbf{Method}}} & \multirow{3}[4]{*}{\textbf{\# Para}} & \multicolumn{5}{c}{\textbf{Mathematical Reasoning}} & \multicolumn{3}{c}{\textbf{Complex QA Reasoning}} \\
\cmidrule(lr){3-7} \cmidrule(lr){8-10}
&       & \textbf{GSM8K} & \textbf{G\_Hard} & \textbf{SVAMP} & \textbf{ASDIV} &  \textbf{MultiArith} & \textbf{MuSiQue} & \textbf{HotpotQA} \\
\midrule
\headercolor
\multicolumn{9}{c}{\textbf{Using 1.3B Local Agent}} \\
Random Switch & 1.3B    & 36.50  & 29.90  & 35.60  & 50.60  & 85.56  & 31.00 & 24.80\\
Sequential Switch & 1.3B & 36.00  & 29.90  & 31.80  & 47.90  & 82.78 & 26.80 & 28.80\\
Confidence Switch & 1.3B & 38.70  & 33.40  & 35.60  & 53.40  & 89.44 & 29.80 & 29.50 \\
\midrule
\methodname{}& 1.3B    & 53.90  & 47.10  & 46.90   & 61.90 & 94.44  & 36.80 &  32.50 \\
\midrule
\headercolor
\multicolumn{9}{c}{\textbf{Using 3B Local Agent}} \\
Random Switch & 3B      & 54.40  & 49.80  & 45.40  & 60.40  & 94.44  & 35.50 & 26.80\\
Sequential Switch & 3B  & 52.70  & 48.80  & 43.00  & 60.40  & 93.33  & 34.50 & 30.50\\
Confidence Switch & 3B  & 54.70  & 48.90  & 47.60  & 63.20  & 95.00  & 35.80 & 31.00\\
\midrule
\methodname{} & 3B   & 60.60  & 50.60  & 52.60 & 66.20  & 96.11  & 37.80 & 31.80 \\
\bottomrule
\end{tabular}
}
\caption{Results (\%) of different switching strategies. We conduct the experiment by using 1.3B and 3B as the local agents to dynamically activate the 33B cloud agent. }
\label{tab:switching}%
\end{table*}%

\paragraph{Generalization Ability}
In this section, we aim to analyze whether the local agent can collaborate with different cloud agents without further retraining.
Specifically, we utilize both the 1.3B and 3B local agents trained under the supervision of the 33B cloud agent and ask the local agent to ask for help from other cloud agents when it thinks it has made a mistake.
We choose cloud agents of different parameter sizes, which include 13B, 32B, and 70B. Based on the result shown in \cref{tab:general}, we can have the following conclusions:

First, when switching to unknown cloud agents, the local agent's performance can still be improved significantly, which demonstrates the generalization ability of our method.
Moreover, the improvement ratio is higher when switching to larger cloud agents with better model capability.
For example, the relative improvement of \ours{} achieves surprising 101\% and 109\% on GSM8K and SVAMP when the 1.3B local agent collaborates with the 70B cloud agent.
This is important because we can only deploy one local agent on the local device while deploying multiple cloud agents remotely.
During inference, the users can dynamically decide which cloud agents to use based on the quota of computational resources.

\paragraph{Cost-Effectiveness Analysis}
In this section, we analyze the cost and effectiveness of existing methods.
Specifically, we take GSM8K as our test bed and calculate the average cost per query utilizing different inference modes, such as Random Switch, Sequential Switch, and Confidence Switch.

\begin{figure}[tb]
\includegraphics[width=0.95\linewidth]{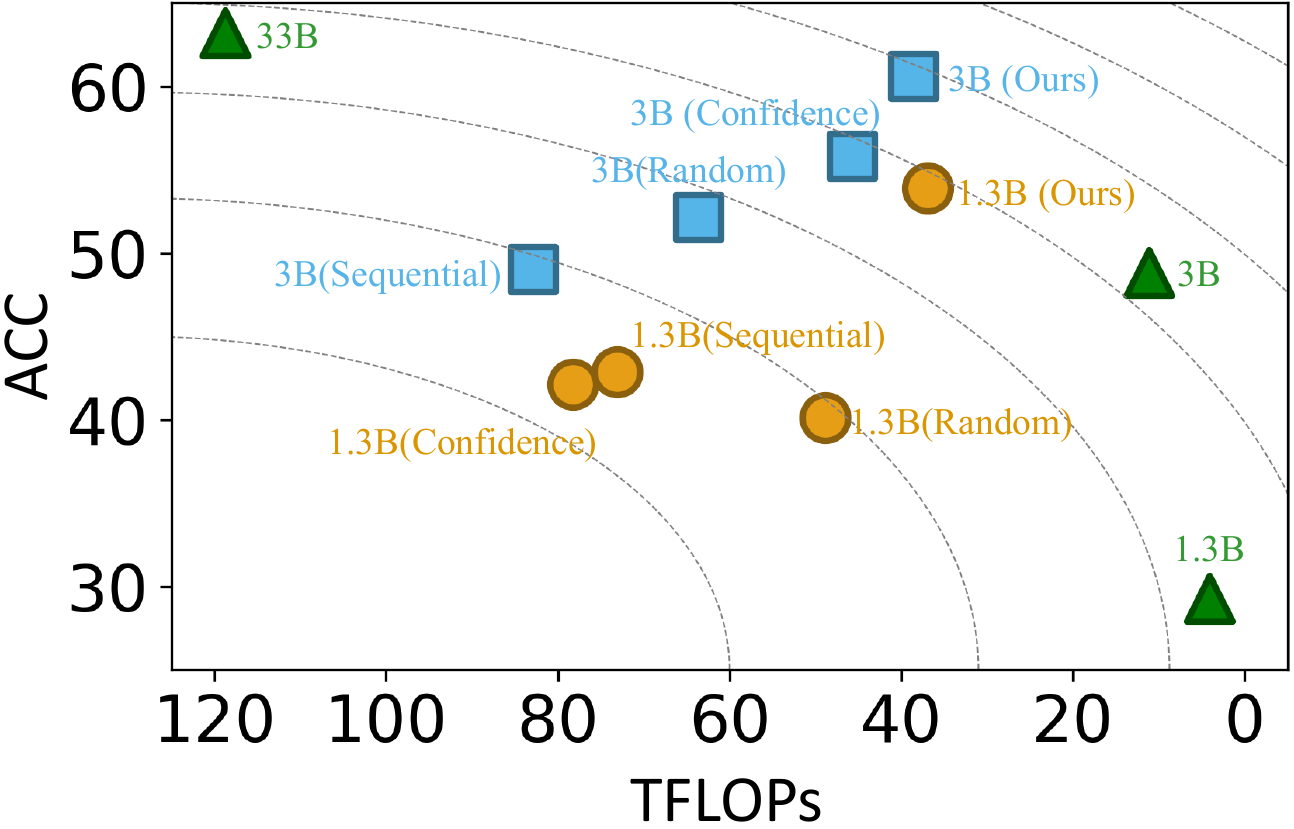}
\centering
\caption{Cost-Effectiveness Analysis. We conduct experiments on the GSM8K dataset. From left to right, the cost of the methods gradually increases. From the bottom to the top, the accuracy of the method increases.}
\label{fig:cost_analysis}
\end{figure}

As shown in \cref{fig:cost_analysis}, the 33B local agent achieves the highest performance with the largest computational cost, and the 1.3B local agent achieves the lowest performance with the lowest cost.
As a comparison, \ours{} balances cost and effectiveness and achieves a similar cost with 33B agent with 3x fewer computational costs.

\subsection{Case Study}
\begin{figure*}[tb]
\includegraphics[width=0.95\textwidth]{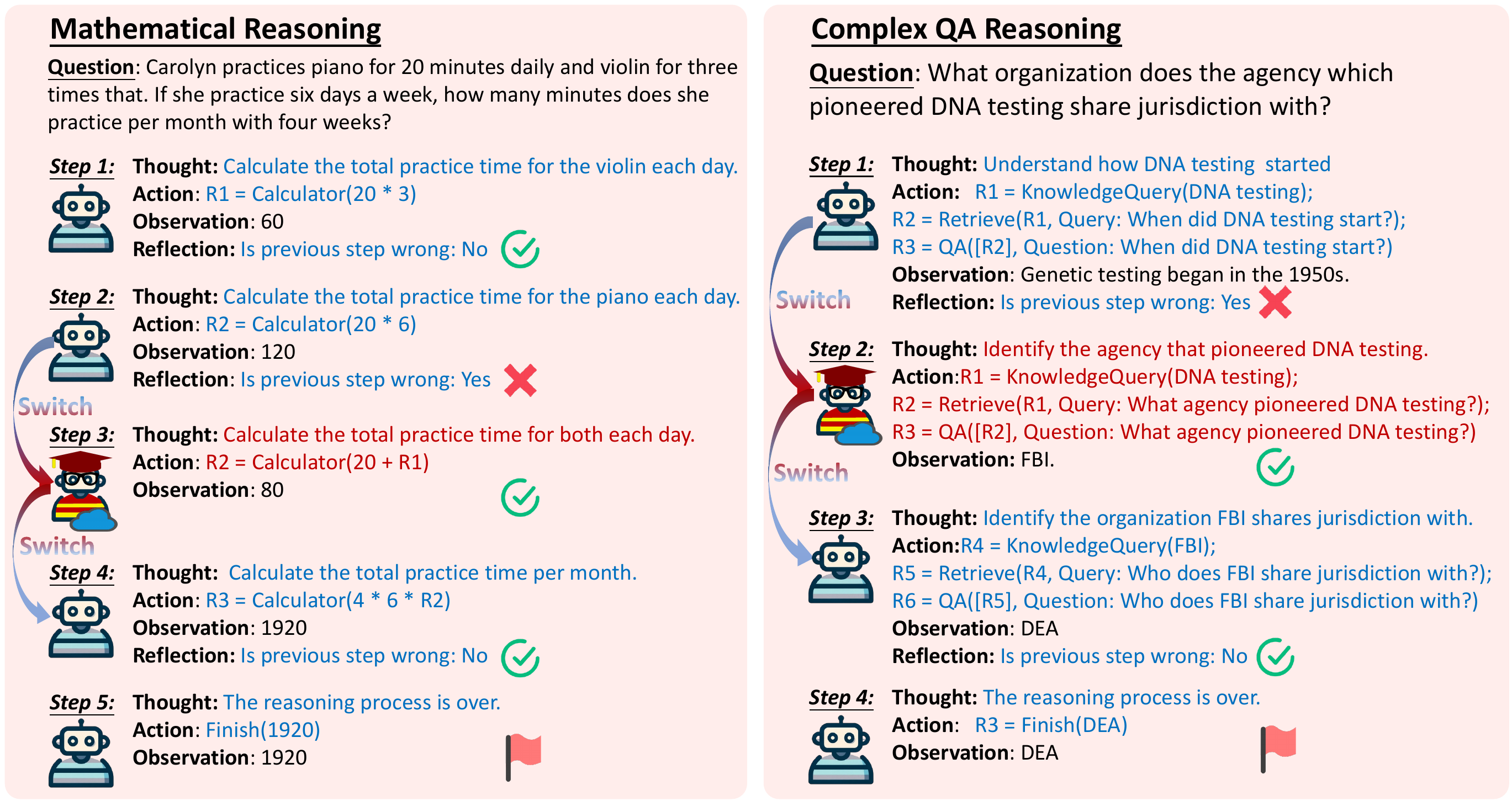}
\centering
\caption{Case studies of solving Mathematical  Reasoning and Complex QA Reasoning problems. Blue text indicates the generation from the local agent while Red text indicates the generation from the cloud agent}
\label{fig:confusion}
\end{figure*}
In this section, we examine examples from Mathematical Reasoning and Complex QA Reasoning tasks to illustrate the performance of \ours{}.

For instance, in a mathematical reasoning task involving a question about piano practice time, the local agent correctly calculates the total daily practice time for the violin but fails to compute the total practice time for the piano. However, due to its strong reflection capability, the local agent recognizes the error. At this point, a robust cloud agent intervenes and successfully corrects the mistake. The local agent then resumes the generation process based on the corrected input.
Similarly, in a complex QA reasoning task, the local agent initially fails to decompose the questions correctly but soon realizes the mistake. The cloud agent intervenes, accurately predicting the next step, and allowing the local agent to continue the inference process until it reaches the correct answer.
Through these examples, we can find that the local agent can successfully finish most easy steps while needing intervention from the cloud agent in rare cases, making \ours{} meaningful and necessary.
\section{Related Work}

\subsection{Multi-Agent Collaboration}

Enabling multiple models with different characteristics to collaborate can lead to stronger performance or lower computational costs than a single model. $\alpha$-UMi \cite{shen2024small} divides a tool learning task that is challenging for a single small LLM into three subtasks, which are delegated to three small LLMs to collaborate on. In some scenarios, choosing the appropriate agent presents a challenge, and Co-LLM \cite{shen2024learning} achieves token-level agent selection during inference by learning latent variables. AutoMix \cite{madaan2023automix} uses an external meta-verifier to assess the correctness of small model outputs and decides whether to route to a large model, achieving query-level agent selection. SwiftSage \cite{lin2024swiftsage} makes rule-based judgments based on feedback from the external environment to determine whether to switch from Fast mode to Slow mode, achieving action-level agent selection. Corex \cite{sun2023corex} introduces a suite of strategies designed to enhance the capabilities of LLMs in complex task-solving, with a pivotal focus on advancing multi-model collaboration.

In \methodname{}, through self-reflection of local agents and spontaneous collaboration with cloud agents, achieves a better balance between cost and performance.

\subsection{Learning from Mistakes}
The methods of learning from mistakes are mainly categorized into two types: prompt-based and finetune-based. For prompt-based methods, 
TRAN \cite{yang2023failures} summarizes the reasons for past errors of the LLM as rules, forming a set of rules. During the inference stage, the model retrieves rules from the rule set as part of the prompt to assist in model reasoning. LEAP \cite{zhang2024context} improves the few-shot prompt by intentionally making the model make mistakes when solving few-shot examples, allowing the model to reflect on errors and acquire task-specific principles, which help prevent making similar mistakes in the future.
RICP \cite{sun2024retrieved} proposes to retrieve the relevant insights from previous mistakes and apply hierarchical clustering to the reasons and insights.

For finetune-based methods, LEMA \cite{an2023learning} corrects model errors with GPT-4 and uses the correction process as a new dataset for the model to learn self-correction. By adding positive and negative prefixes to correct and incorrect rationals in the training data, mistake tuning \cite{tong2024can} and NAT \cite{wang2024learning} can enhance model performance in the inference stage using positive prefixes. \citet{wang2023democratizing} proposed method enhances the model's reasoning ability by allowing the model to learn from self-reflection and customized feedback.

In \methodname{}, we utilize a finetune-based approach to enable the model to self-assess errors, thereby allowing for self-correction or switching to a more powerful LLM for assistance.
\section{Conclusion}

In this work, we advocate \methodname{}, a novel multi-agent collaboration framework that effectively integrates the strengths of both locally deployed and cloud-based LLMs.
The local agent is responsible for less complex reasoning steps, and the cloud agent is dedicated to intricate reasoning steps.
Experimental results and in-depth analysis demonstrate that \methodname{} is able to bring significant improvements in task performance while using much less computational overhead.

\section*{Limitations}

In this work, we evaluate our proposed framework on mathematical reasoning and complex question answering tasks, and it remains to be investigated in future work how to extend our approach to a wider range of reasoning tasks.
Besides, due to constraints on computational resources and funding, we do not conduct experiments on larger scale language models (>100B). Thus the performance of larger LLMs remains undetermined.
We will further explore the performance of our framework on larger scale language models in future research.

\section*{Acknowledgement}
This work is supported in part by Ucap Cloud and the State Key Laboratory of General Artificial Intelligence.
\section*{Ethics Statement}
This work was conducted in strict compliance with the ACL Ethics Policy. 
All datasets and large language models (LLMs) used for evaluation are publicly available. 
Furthermore, our work aims to explore a multi-agent collaboration framework. 
We do not foresee any negative ethical impacts arising from our work.

\bibliography{custom}
\clearpage
\appendix
\onecolumn
\section{Dataset Statistics}

The dataset statistics used in this paper is shown in Table~\ref{tab:datasets}.

\section{Tool Definition}
The tool definition is listed in Table~\ref{table:appendix-tools}.

\begin{table*}[htbp]
\centering
\resizebox{\linewidth}{!}{
\begin{tabular}{l|cccccccc}
\toprule
\multicolumn{1}{c|}{\multirow{2}{*}{\textbf{Method}}} & \multicolumn{5}{c}{\textbf{Mathematical Reasoning}} & \multicolumn{3}{c}{\textbf{Complex QA Reasoning}} \\
\cmidrule(lr){2-6} \cmidrule(lr){7-9}
 & \textbf{GSM8K} & \textbf{G\_Hard} & \textbf{SVAMP} & \textbf{ASDIV} &  \textbf{MultiArith} & \textbf{MuSiQue} & \textbf{HotpotQA} & \\
\midrule
Train Data & \multicolumn{5}{c}{7,500 instances from GSM8K}  & \multicolumn{2}{c}{10,000 instances from MuSiQue} \\
Test Data & 1,000 & 1,000 & 1,000 & 1,000 & 1,000 & 500 & 500 & \\
\bottomrule
\end{tabular}%
}
\centering
\caption{Datasets Statistics.}
\label{tab:datasets}
\end{table*}
\begin{table}[htbp]
\centering
\begin{subtable}[t]{\textwidth}
\centering
\scalebox{0.7}{
\begin{tabular}{cccc}
\toprule
\textbf{Task Type}                  & \textbf{Action Types}   & \textbf{Function Descriptions} & \textbf{Tools} \\ \midrule
\multirow{7}{*}{QA} & \texttt{KnowledgeQuery(Entity) -> Knowledge}    & Query the entity knowledge   & Wikipedia, Google Search              \\ \cmidrule{2-4}
& \makecell{\texttt{ParagraphRetrieval(Knowledge, Query)} \\ \texttt{-> Paragraphs}}    & \makecell{Retrieve relevant paragraphs \\ according to the query}   & \texttt{dpr-reader-multiset-base}              \\ \cmidrule{2-4}
& \texttt{QA(Context, Query) -> Answer}    & \makecell{Answer the query based on \\ the given context}   & GPT-series/open LLMs              \\ \cmidrule{2-4}
& \texttt{Calculator(Expression) -> Value}
    & Calculate given math expressions   & WolframAlpha \\ \bottomrule
\end{tabular}
}
\caption{Actions used in Complex QA Tasks.}
\label{table:appendix-tools-qa}
\end{subtable}
\vspace{9pt}

\begin{subtable}[t]{\textwidth}
\centering
\scalebox{0.7}{
\begin{tabular}{cccc}
\toprule
\textbf{Task Type}                  & \textbf{Action Types}   & \textbf{Function Descriptions} & \textbf{Implementation} \\ \midrule
\multirow{8}{*}{Math}      & \texttt{Calculator(Expression) -> Value}
& Calculate given math expressions   & \multirow{6}{*}{WolframAlpha}     \\ \cmidrule{2-3}
& \texttt{SetEquation(Expression) -> Equation} & Set equations based on given expressions &                    \\ \cmidrule{2-3}
& \texttt{SolveEquation(Equation) -> Solutions}      & Solve the set equations  &                  \\ \cmidrule{2-3}
& \texttt{Define(Variable) -> Variable} & Define a variable &                    \\ \cmidrule{2-3}
& \texttt{SolveInequality(Inequality) -> Solutions}      & Solve the given inequality  &                  \\ \cmidrule{2-4}
& \texttt{Code(Function\_Description) -> Code} & Generate codes for math functions & \texttt{gpt-3.5-turbo}        \\ \cmidrule{2-4}
& \texttt{Count(List) -> Number}      & Count the element number in a list  & Python                 \\ \bottomrule
\end{tabular}
}
\caption{Actions used in Mathematical Tasks.}
\label{table:appendix-tools-math}
\end{subtable}
\vspace{9pt}

\caption{Action interfaces and execution module used in this paper.}
\label{table:appendix-tools}
\end{table}

\end{document}